\documentclass[a4paper]{article}
\usepackage{ISCSLP2026}
\usepackage{ifthen}
\usepackage{url}
\newboolean{blind}
\setboolean{blind}{false} 
\title{Probing in the Wild: A Case Study of Self-Supervised Speech Representations on Mandarin Sub-dialects with Unsupervised Articulatory Analysis}

\name{
	\ifthenelse{\boolean{blind}}{Anonymous to ISCSLP}
	{Shu Shang$^1$, Fuliang Weng$^2$, Zeqian Hu$^1$, Yaqian Zhou$^1$}
}

\address{
	\ifthenelse{\boolean{blind}}{Anonymous to ISCSLP}
	{
		$^1$ Fudan University, China \\
		$^2$ Logos \& Dialogos, US
	}
}

\email{
	\ifthenelse{\boolean{blind}}{Anonymous to ISCSLP}
	{sshang23@m.fudan.edu.cn, fuliang@gmail.com, zqhu24@m.fudan.edu.cn, zhouyaqian@fudan.edu.cn}
}

\newcommand{\indexterms}{interpretability, self-supervised learning, articulatory features, dialect variation}
\begin{document}
	
	\maketitle
	
	\begin{abstract}
		While self-supervised speech models have achieved strong performance across speech tasks, relatively little is known about how their internal phonetic representations behave under fine-grained dialect variation. Existing probing studies typically rely on curated corpora with manual phonetic annotations, limiting their applicability to naturally occurring dialect speech. We present a case study of articulatory feature representations in a Mandarin self-supervised speech model using an entirely unlabeled probing pipeline. Phone sequences are generated using a language-agnostic universal phone recognizer and mapped to articulatory feature vectors, enabling frame-level probing without manual annotation. Our results reveal a structured pattern in articulatory feature decodability across Mandarin sub-dialects. Acoustically salient features such as labiality and stridency remain comparatively stable, whereas features associated with finer spectral distinctions exhibit larger dialect-dependent variation. This variation is driven primarily by elevated decodability for Beijing speech relative to other Mandarin sub-dialects. Layer-wise analyses further show distinct representational dynamics for these feature groups. These findings suggest that language-agnostic articulatory probing can be applied to real-world dialect corpora and that dialect sensitivity in self-supervised speech representations is unevenly distributed across articulatory dimensions.
	\end{abstract}
	
	\section{Introduction}
	Self-supervised speech representation models have become the default backbone for modern speech systems by enabling models to learn general-purpose acoustic representations from unlabeled audio, which are then transferred to downstream tasks. However, they remain opaque in how they encode and represent the wealth of phonetic information within the audio data. 
	
	To fill this gap, prior work has explored how speech models organize phonemic information \cite{mohamed2022}. Using methods such as probing classification (e.g. \cite{cormacenglish2022, martin2023, mohamed2024, belinkov2017analyzing, yang21c_interspeech}), feature attribution (e.g. \cite{fucci_echoes_2025,markert21_spsc}), and correlation analysis (e.g. \cite{pasad2024, chung2021, abdullah21_interspeech, silfverberg-etal-2018-sound}). While these studies provide valuable insights into the phonetic structure of learned representations, they are typically conducted with curated datasets. In particular, analyses are frequently conducted on phonetically annotated corpora such as TIMIT \cite{garofolo1993darpa}, which provides a rich phonetic inventory and abundant annotation. This, however, does not reflect the acoustic and sociolinguistic variability in real-life application. As such, existing work largely characterizes how SSL models encode canonical, standard pronunciations, rather than how they generalize to dialectal and sub-dialectal diversities. 
	
	Chinese provides a particularly good testbench for this question because its linguistic variations appear in a hierarchy. In Chinese linguistics, the Sinitic language family is typically divided into several major branches (e.g., Mandarin, Wu, Yue, Min). Many of these branches are mutually unintelligible even as they are commonly referred to as ``dialects'' in existing literature. In contrast, within Mandarin (Guanhua) itself, there exist multiple regional sub-dialects, such as Jiang-Huai and Southwestern Mandarin, that share a common phonological backbone but differ in fine-grained phonetic details \cite[pp.~72--83]{norman2003chinese}. In this paper, “Mandarin sub-dialects” refers to regional varieties within the Mandarin/Guanhua branch, not to broad Sinitic branches such as Wu, Yue, or Min, and not merely to accent variation around Standard Mandarin.
	
\begin{figure}[th]
		\centering
		\includegraphics[width=\linewidth,trim=1pt 1pt 1pt 1pt, clip]{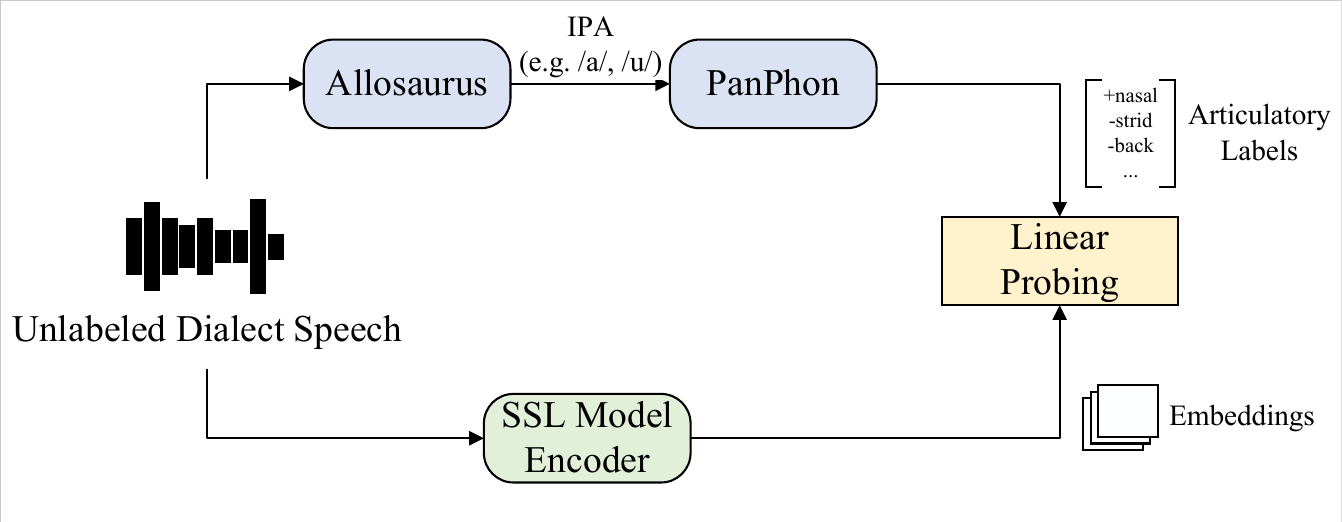}
		\caption{Overview of our unsupervised probing pipeline.}
		\label{fig:overview}
	\end{figure}

	Prior cross-lingual and cross-dialect work usually focus on distinct language or accent contrasts \cite{deseyssel22_interspeech}, but does not elaborate on the robustness of representations under subtle phonetic shifts. A key obstacle in studying such fine-grained variations is the lack of phonetic annotation at scale. Unlike curated corpora, real-world speech datasets rarely provide frame-level phoneme or articulatory labels, making conventional probing approaches difficult to apply. To overcome this limitation, we adapt an unsupervised probing pipeline (shown in Figure~\ref{fig:overview}) that uses language-agnostic pseudo-labeling to obtain articulatory feature representations. Speech is first converted into phone sequences using a universal phone recognizer and then mapped into articulatory feature representations. Using these pseudo-labels, we probe frozen self-supervised representations trained on Mandarin and examine the robustness of articulatory features across sub-dialects. Rather than asking whether phonetic information exists in the model, our goal is to determine whether that information remains stable under naturally occurring cross-dialectal variation.

	In summary, our contributions are:
	\begin{itemize}
		\item We adapt a language-agnostic pseudo-labeling pipeline for articulatory probing of unlabeled Mandarin sub-dialects.
		\item We show that articulatory features exhibit a structured hierarchy in cross-dialect linear decodability.
		\item We identify instability in fine-grained spectral features, revealing dialect-specific discrepancies in a pretrained wav2vec 2.0 \cite{baevski2020} model.
	\end{itemize}
	\section{Methods}
	\subsection{Dataset}
	To study how SSL representations encode fine-grained Mandarin sub-dialect variation, we require a corpus containing multiple closely related sub-dialects with sufficient speaker diversity and consistent recording conditions. 
	
	We conduct our experiments on the KeSpeech corpus \cite{tang2021kespeech}, a large-scale Mandarin speech dataset containing recordings from eight Mandarin sub-dialects spanning distinct geographic regions. This diversity makes this corpus well suited for studying fine-grained sub-dialect discrimination. In the KeSpeech sub-dialect identification benchmark, recordings are annotated with nine labels: eight regional Mandarin sub-dialects and an additional ``mandarin'' category, corresponding to examples judged by annotators to be close enough to classify as standard Mandarin. For the sake of category homogeneity, we only use the eight sub-dialect labeled examples, and do not consider the ``mandarin'' label. We adopt the predefined ``test'' split from the KeSpeech sub-dialect identification task. For each sub-dialect, we balance speaker counts and construct non-overlapping train/test partitions at the speaker level to prevent speaker leakage. After preprocessing, the resulting dataset comprises approximately 3 hours of speech across eight Mandarin sub-dialects. Because the probing task operates at the frame level, the dataset yields approximately 80,000 labeled frames, providing sufficient training instances for linear probing.
	
	\subsection{Model}
	Our analysis pipeline consists of two components: a language-agnostic articulatory labeling pipeline that converts speech waveforms into frame-aligned articulatory feature targets, and a frozen self-supervised speech encoder from which we extract intermediate layer representations. Linear probes are trained on the representation to obtain the model's perception of articulatory features.
	
	\subsubsection{Articulatory Feature Extraction}
	We obtain frame-aligned International Phonetic Alphabet (IPA) phone sequences from raw waveforms using the Allosaurus universal phone recognizer \cite{li_universal_2020}. Allosaurus is configured to operate in \texttt{ipa} mode, which produces language-agnostic phone predictions rather than constraining outputs to a predefined language inventory. To synchronize articulatory supervision with the SSL encoder, we use timestamp generated directly by Allosaurus using its  connectionist temporal classification (CTC) decoder, and quantize it to the nearest wav2vec 2.0 frame center. The resulting IPA sequences are mapped to ternary articulatory feature vectors using PanPhon \cite{mortensen_panphon_2016}, which deterministically converts each IPA segment into a vector of 22 distinctive articulatory features encoded in a ternary scheme (positive, negative, or unspecified). \footnote{The PanPhon feature inventory includes the feature \textit{velaric} (velaric airstream mechanism, click), which is not meaningfully present in Mandarin. We retain its statistics for completeness but exclude it from subsequent analysis.}
	
	Crucially, this procedure requires no manual phonetic annotation, allowing probing analysis on naturally occurring speech corpora that would otherwise be unusable for articulatory evaluation. Together, the Allosaurus-PanPhon pipeline provides frame-level articulatory targets that are both language-agnostic and physically interpretable, serving as pseudo-label supervision for subsequent probing.
	
	\subsubsection{Self-supervised Speech Encoder}
	We use a pretrained wav2vec 2.0 encoder as the source of self-supervised speech representations. Specifically, we adopt the \texttt{chinese-wav2vec2-base} checkpoint\footnote{\url{https://huggingface.co/TencentGameMate/chinese-wav2vec2-base}}. This is a SSL model with 95M parameters trained on WenetSpeech \cite{zhang2022}, a large-scale Mandarin speech corpus curated from YouTube and Podcast data. This model is representative of modern SSL encoders used in real-life Mandarin speech tasks.
	
	We focus on embeddings from an intermediate transformer layer, consistent with prior work indicating that middle layers retain strong phonetic information \cite{pasad2021}. Layer 7 is selected for detailed feature-level analysis because the hierarchy is most clearly expressed at intermediate depths. 
	
	\subsection{Linear Probing}
	We conduct articulatory probing as a frame-level classification task on frozen SSL embeddings.
	
	We probe the contextualized hidden state at the Allosaurus-aligned frame center without additional temporal pooling. Thus, any temporal information available to the probe must already be encoded in the SSL model’s contextualized representation rather than introduced by the probing window. All embeddings are standardized using feature-wise z-score normalization computed from the training set statistics prior to probe training. Independent linear probes are trained for each articulatory feature and encoder layer using multinomial logistic regression with the \texttt{qn} solver from the cuML package \cite{raschka2020machine}. To prevent speaker leakage, rigorous group-based partitioning is used to ensure speaker-independent train/test splitting. 
	
	Performance is evaluated using frame-level Macro-F1 score. For each encoder layer and articulatory feature, Macro-F1 scores are computed to quantify the extent to which the corresponding articulatory feature is linearly decodable from the learned representations.
	\section{Results}
	\subsection{Heterogeneity of Representation}
	\begin{table*}[th]
		\centering
		\caption{Macro-F1 scores of articulatory features sorted by the absolute Beijing--others gap $|\Delta|$. $\text{Avg}$ indicates the arithmetic mean of the Macro-F1 scores of the other seven sub-dialects, $\Delta=\text{Avg}-\text{Beijing}$. The ordering reveals a visible discontinuity in $|\Delta|$ values, with a substantial margin between the larger- and smaller-gap features. Features in bold correspond to the higher-divergence group.}
		%	\resizebox{\linewidth}{!}{%
			\begin{tabular}{lrrr|lrrr}
				\toprule
				\multicolumn{4}{c}{\textbf{High-gap}} &
				\multicolumn{4}{c}{\textbf{Low-gap}} \\
				Feature & Avg & Beijing & $\Delta$ &
				Feature & Avg & Beijing & $\Delta$ \\
				\midrule
				\textbf{velaric}              & 0.5406 & 0.8331 & \textbf{-0.2924} &
				spread glottis               & 0.6252 & 0.6589 & -0.0337 \\
				\textbf{long}                 & 0.3966 & 0.6267 & \textbf{-0.2301} &
				low                          & 0.5720 & 0.6037 & -0.0317 \\
				\textbf{syllabic}             & 0.6158 & 0.8246 & \textbf{-0.2088} &
				tense                        & 0.6374 & 0.6631 & -0.0256 \\
				\textbf{coronal}              & 0.5948 & 0.7979 & \textbf{-0.2032} &
				anterior                     & 0.7141 & 0.7358 & -0.0218 \\
				\textbf{nasal}                & 0.5819 & 0.7818 & \textbf{-0.1998} &
				high                         & 0.6559 & 0.6771 & -0.0212 \\
				\textbf{consonantal}          & 0.6189 & 0.8181 & \textbf{-0.1992} &
				delayed release              & 0.5793 & 0.5583 & 0.0211 \\
				\textbf{back}                 & 0.5932 & 0.7832 & \textbf{-0.1900} &
				distributed                  & 0.6147 & 0.6347 & -0.0200 \\
				\textbf{sonorant}             & 0.6259 & 0.8159 & \textbf{-0.1900} &
				labial                       & 0.6409 & 0.6529 & -0.0120 \\
				\textbf{lateral}              & 0.5010 & 0.6477 & \textbf{-0.1467} &
				continuant                   & 0.6975 & 0.7039 & -0.0064 \\
				\textbf{constricted glottis}  & 0.4163 & 0.5552 & \textbf{-0.1389} &
				strident                     & 0.6590 & 0.6641 & -0.0050 \\
				\textbf{voice}                & 0.6112 & 0.7478 & \textbf{-0.1366} &
				round                        & 0.6150 & 0.6179 & -0.0028 \\
				\bottomrule
			\end{tabular}%
			%	}
		
		\label{tab:gap_sorted}
	\end{table*}
	
\begin{figure}[th]
		\centering
		\includegraphics[
		width=\linewidth
		]{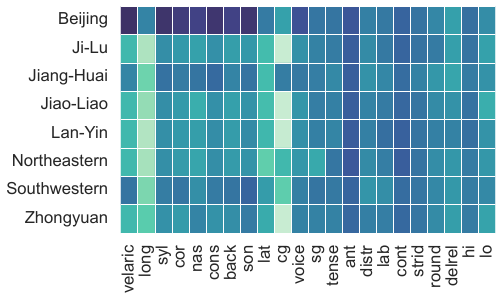}
		\caption{Heatmap of Macro-F1 scores for articulatory feature decoding across eight Mandarin sub-dialects. Darker colors indicate higher classification performance.}
		\label{fig:perdialectheatmap}
	\end{figure}

	Figure \ref{fig:perdialectheatmap} shows a per-dialect and per-feature breakdown of the model's articulatory representation ability. A clear boundary is drawn between the Beijing Mandarin sub-dialect and all other Mandarin sub-dialects.
	
	As a qualitative sanity check, we examine whether the probing results recover known phonological patterns. One such example is the \textit{constricted glottis} feature of Jiang-Huai Mandarin. As shown in Figure \ref{fig:perdialectheatmap}, Jiang-Huai Mandarin obtains the highest Macro-F1 score among all Mandarin variants, including the Beijing sub-dialect. This aligns with the well-known phonological fact that Jiang-Huai Mandarin is the only Mandarin sub-dialect that systematically preserves the ``Entering Tone'', a tone characterized by its short duration and a final glottal stop.
	\subsection{Hierarchy of Representation}
	\begin{figure}[th]
		\centering
		\includegraphics[width=\linewidth]{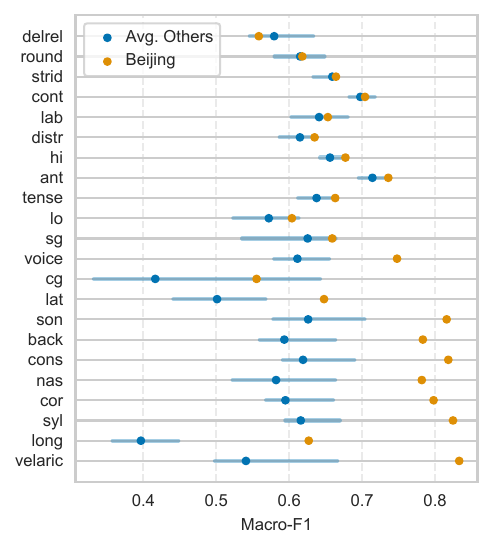}
		\caption{Comparison of Beijing speech to the envelope of seven other Mandarin sub-dialects across phonological features. For each feature, the horizontal bar shows the range of scores among the other dialects, the blue point indicates their mean, and the orange point indicates Beijing.}
		\label{fig:dumbbell}
	\end{figure}
	
	The disparity between the Beijing sub-dialect and other varieties is further visualized in Figure \ref{fig:dumbbell}. Table~\ref{tab:gap_sorted} summarizes this pattern by sorting features according to the absolute Beijing--others gap $|\Delta|$. Features with small gaps ($|\Delta|<0.10$) exhibit stable decoding performance across dialects, whereas a subset of features shows pronounced Beijing-centric discrepancies.
	
	We observe a clear hierarchy regarding the robustness of different articulatory features. Features such as \textit{strident}, \textit{anterior}, \textit{labial}, and \textit{high} exhibit remarkable stability across dialects. These features share a common physical property: they are acoustically salient and structurally stable. In contrast, features including \textit{nasal}, \textit{back}, and \textit{coronal} show substantial disparity between Beijing Mandarin and other sub-dialects. These features also share a common physical property: they are acoustically subtle and more sensitive to contextual or coarticulatory variation. Notably, this high disparity is driven exclusively by an elevated performance on the Beijing sub-dialect, rather than a collapse on other sub-dialects.
	
	\subsection{Layer-wise Analysis}
	\begin{figure}[th!]
		\centering
		\includegraphics[width=\linewidth]{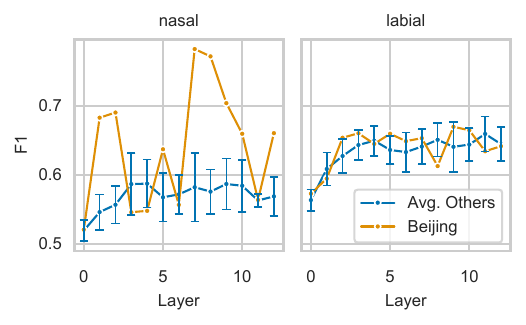}
		\caption{Layer-wise Macro-F1 scores for a stable feature \emph{(labial)} and an unstable feature \emph{(nasal)} comparing Beijing to the average of other Mandarin sub-dialects. Points indicate mean F1 per layer. Error bars represent standard deviation across non-Beijing sub-dialects.}
		\label{fig:perlayer}
	\end{figure}
	To examine the representational dynamics across network depth, we track the probing performance of the representation from the initial convolutional head output (layer 0) through all 12 transformer layers. Figure~\ref{fig:perlayer} contrasts the layer-wise performance shifts of a stably linearly decodable feature (\emph{labial}) and an unstable feature (\emph{nasal}). These features are representative of their respective groups; other features display qualitatively similar layer-wise variability.
	
	The stable feature demonstrates early convergence. The performance for both Beijing sub-dialect and other sub-dialects plateau around layer 3, and remain stable throughout the network. This confirms that acoustically salient features are well-represented early on. Conversely, the unstable feature exhibits extreme layer-wise instability. While the performance for other sub-dialects follow roughly the same pattern as robust features, performance for the Beijing sub-dialect is highly erratic. We observe dramatic performance spikes and sharp degradations, reinforcing our finding that the model fails to find a stable and robust abstraction for such features.
	
	\section{Discussion}
	\subsection{Hierarchy in Sensitivities to Acoustic Features}
	An examination of the probing performance reveals an asymmetry in how SSL models generalize in different acoustic properties. For robustly encoded features (e.g. \emph{strident}, \emph{labial}), the model maintains a consistent baseline across dialects. However, fine-grained features (e.g. \emph{back}, \emph{coronal}) exhibit significantly higher sensitivity to dialectal variations.
	
	The model shows an anomalously high performance on the standard Beijing sub-dialect, leading to a large performance gap between Beijing and other 7 sub-dialects. This performance gap suggests that the encoding of subtle spectral shifts is more tightly coupled with the specific acoustic distribution on the pretraining data. When evaluated on sub-dialects that naturally deviate from these standard distributions, the representations become less linearly separable, reflecting a specific vulnerability in modeling continuous phonetic variations.

	\begin{table}[th]
		\caption{Representative mapping of standard phonemes to articulatory features. `+' and `-' denote positive and negative specification of the feature.}
		\label{tab:phon}
		\centering
		\begin{tabular}{c|cc|ccc}
			\toprule
			\textbf{Phoneme} & \textbf{labial} & \textbf{strident} & \textbf{coronal} & \textbf{nasal} & \textbf{back} \\
			\midrule
			\textbf{/b/} & + & - & - & - & - \\
			\textbf{/m/} & + & - & - & + & - \\
			\textbf{/s/} & - & + & + & - & - \\
			\textbf{/a/} & - & - & - & - & + \\
			\textbf{/i/} & - & - & - & - & - \\
			\textbf{/u/} & + & - & - & - & + \\
			\bottomrule
		\end{tabular}
	\end{table}
	
	Table~\ref{tab:phon} shows the deterministic mapping from standard phonemes to their articulatory features using the PanPhon database. This mapping explicitly translates abstract phonemic categories into physical properties.

	For example, the robust \emph{labial} feature relies on absolute occlusions, as seen in /p/ and /m/. The vulnerable \emph{back} feature captures the continuous posterior displacement of the tongue body, as seen in /a/ and /u/. By decomposing phoneme into articulatorially meaningful vectors, we allow our probing process to directly evaluate the model's capacity in encoding specific physical articulatory organs.
	
	This feature-level decomposition allows us to see the diverse nature of phonemes. Phonemes are commonly used as the smallest unit in prior research. However, we demonstrate that many phonemes are conglomerates of stable and unstable features. The /u/ phoneme, for instance, shows both \emph{labial}, a robust feature, and \emph{back}, an unstable feature. 
	\subsection{Validity Under Noisy Supervision} 
	A potential confounding factor in our work is the reliability of the pseudo-labels generated by the Allosaurus-PanPhon pipeline. The observed performance gap ($\Delta$) between the Beijing sub-dialect and other sub-dialects may reflect the degraded Phone Error Rate of the annotator, rather than the representational nuances of the encoder.
	
	However, if this were the primary driver, we would expect relatively uniform degradation across articulatory dimensions or inconsistent feature-level patterns. Instead, we observe a structured hierarchy in which only a subset of fine-grained features exhibits substantial cross-dialect divergence, while others remain stable. Moreover, the probe captures dialect-specific phonological traits (e.g., constricted glottis in Jiang-Huai Mandarin), suggesting that the pseudo-labeling pipeline retains linguistically meaningful signal. 
	
	\section{Limitations and Future Directions}
	Despite these findings, our approach has several limitations. First, linear probing only measures the linear decodability of representations. This may not accurately reflect how such representations are utilized by downstream tasks. Second, our current work examines only a single SSL model (wav2vec 2.0) trained on Mandarin speech and evaluated on Mandarin sub-dialects. Extending this pipeline to other architectures and language families is necessary to determine whether the observed heterogeneity and hierarchy of representation generalize beyond this case study. Third, our reliance on Allosaurus introduces a potential confounding factor, as the Phone Error Rate of the pseudo-labeler may unevenly degrade across sub-dialects. While our phonological case studies (e.g., Jiang-Huai) suggest that genuine linguistic signals are being recovered, future work should incorporate small-scale manually annotated dialect datasets to better disentangle representational limitations of the SSL encoder from bias in the pseudo-labeling pipeline.
	
	\section{Conclusion}
	This work utilizes a cross-dialectal unsupervised articulatory probing framework to dissect the phonetic representations of self-supervised speech models in unlabeled datasets. Our empirical analysis suggests a potential hierarchy in how SSL encoders represent articulatory features: while SSL encoders successfully generalize in robust acoustic features, it suffers from dialect-specific overfitting on fine-grained features. By identifying this bottleneck, our findings provide insights for the inclusiveness and dialect awareness of speech models.
	\newpage
	
	\bibliographystyle{IEEEtran}
	\bibliography{mybib}
	
\end{document}